\documentclass{article}
\usepackage[utf8]{inputenc}
 \pdfoutput=1 

\usepackage{aaai201}

\usepackage{times}
\usepackage{helvet}
\usepackage{courier}
\usepackage{graphicx}  
\usepackage{url}
\usepackage{xcolor}
\usepackage[linesnumbered,ruled,vlined,algo2e]{algorithm2e}
\usepackage{algorithm,algpseudocode}

\usepackage{comment}

\begin{document}

\title{Diversifying Database Activity Monitoring with Bandits}
\makeatletter
    \renewcommand{\copyright@on}{F}
\makeatother

 \author{Hagit Grushka-Cohen,\textsuperscript{1}
Ofer Biller,\textsuperscript{2}
Oded Sofer,\textsuperscript{3}
Lior Rokach,\textsuperscript{1}
Bracha Shapira,\textsuperscript{1}\\
\textsuperscript{1}{Dept. of Software and Information Systems Engineering,Ben-Gurion University of the Negev, Beer Sheva, Israel}\\
\textsuperscript{2}{IBM Guardium Security Division and IBM Cyber Security Center of Excellence,  Beer Sheva, Israel}\\
\textsuperscript{3}{IBM Guardium Security Division, Tel-Aviv, Israel}\\
hgrushka@post.bgu.ac.il,
 ofer.biller@il.ibm.com,
 odedso@il.ibm.com, 
 liorrk@bgu.ac.il,
 bshapira@bgu.ac.il}

\maketitle

\begin{abstract}

\begin{quote}

Database activity monitoring (DAM) systems are commonly used by organizations to protect the organizational data, knowledge and intellectual properties. In order to protect organizations database DAM systems have two main roles, monitoring (documenting activity) and alerting to anomalous activity. Due to high-velocity streams and operating costs, such systems are restricted to examining only a sample of the activity. Current solutions use policies, manually crafted by experts, to decide which transactions to monitor and log. This limits the diversity of the data collected. Bandit algorithms, which use reward functions as the basis for optimization while adding diversity to the recommended set, have gained increased attention in recommendation systems for improving diversity. 

In this work, we redefine the data sampling problem as a special case of the multi-armed bandit (MAB) problem and present a novel algorithm, which combines expert knowledge with random exploration. We analyze the effect of diversity on coverage and downstream event detection tasks using a simulated dataset. In doing so, we find that adding diversity to the sampling using the bandit-based approach works well for this task and maximizing population coverage without decreasing the quality in terms of issuing alerts about events.

\end{quote}
\end{abstract}

\section{Introduction}
Databases are at the heart of organizational IT infrastructure. For data security, privacy protection, and  data leakage prevention,  many organizations monitor database operations in real-time using database activity monitoring (DAM) systems. Such systems are widely used to help implement security policies, and detect attacks and data abuse. The multi-billion dollar  Google-Waymo vs. Uber case \footnote{https://www.newyorker.com/magazine/2018/10/22/did-uber-steal-googles-intellectual-property}, in which  the data collected (and more specifically, data activity logs) was used to document  data theft, provides a good example of DAM system use and the value of such a system to an organization.

With information overload of  hundreds of thousands of transactions per second, database activity monitoring systems cannot process and log all of the activity into log files. Instead of logging all transactions, DAM systems use policies to decide which transactions to save, which are based on rules and users / activity groups defined by the SO during the system setup.

Policy changes require a lot of manual effort, resulting in a situation in which policies rarely change once they are defined. Furthermore, the system primarily monitors users that match  the security officer's (SO) implicit or  explicit preferences very closely, and this restricts them from discovering concept drifts (such as changes in users' roles). 

This static approach  may cause the “filter bubble” phenomenon in which SOs (the users of the DAM system) are trapped in a subspace of options that are too similar to the defined risk profile,thereby losing the ability to explore beyond what they already know. Such filter bubbles are a known issue for recommendation systems  \cite{nguyen2014exploring,chen2019serendipity}.

To address this challenge, we suggest incorporating the concept of diversity from recommendation systems \cite{Matt2014Escaping} into logging policies. Unlike search engines or recommendations, sampling a more diverse group of users is not technically complicated as the user's transactions risk can be aggregated to a single score \cite{grushka2016cyberrank,evina2019enforcing}. However, logging capacity is constrained, and by focusing solely on diversity, undocumented malicious activity in the high risk  group can be missed. A solution must provide a balance between the exploration of activities and users suspected to be of low risk with the exploitation of activities and users suspected to be of high risk. 

The effect of sampling has been studied in the domain of  anomaly detection systems of network traffic monitoring and DAM   \cite{juba2015principled,jadidi2016intelligent,jadidi2015performance,grushka2017sampling}. These studies found that policy based sampling introduces bias to the down stream anomaly detection systems and suggested specific sampling strategies for avoiding bias.

Multi-armed bandits (MABs) are strategies used for policy setting based on sampling for decision-making \cite{agrawal2012analysis,auer2002finite}. To the best of our knowledge, MABs haven't been used so far by an anomaly detection system  to set data collection policy. Since auditing and anomaly detection system can only learn and model the data available to it, the policy determining which transactions are logged has great influence on the ability to detect anomalies, and which anomalies can be detected, and what actions would be possible to audit later. MABs are used for balancing exploration and exploitation in order to maximize a reward in many domains, including reinforcement learning and recommendation systems. In the stochastic multi-armed bandit problem, the goal is to maximize the reward from a slot machine with multiple arms. Various strategies for balancing exploration/exploitation have been studied \cite{agrawal2012analysis}.

In this work we suggest viewing the diversity problem for DAM system sampling strategies as a MAB problem, where the risk of the transactions logged is used as the reward function. 

Unlike the classic MAB problem, in this special case,  the risk distribution of a user is not static: it may  change naturally when a user changes his/her role, or it can change due to hacking or malware, or when an employee has been compromised. Another difference is that in each round we are not sampling one user's activity but sampling multiple users to monitor, \textit{i.e.} multiple arms are pulled in each round (the collector logs all of the transactions for a list of users).

We define a novel variant of the MAB problem to incorporate capacity, pulling multiple levers in each time frame, with a reward function which may change, named Budget-Constrained-Dynamic-MAB.  We present a novel sampling strategy for sampling Budget-Constrained-Dynamic-MAB, a variant of the   $\epsilon$\textendash	Greedy algorithm, named  C\textendash$\epsilon$\textendash	Greedy. 

Using simulated data-sets created with \cite{grushka2019simulating}, we assess the effect of diversity on sampling policies. We introduce two evaluation measures:
(i) coverage of user activity, \textit{i.e.}, how much do we know about each user in the population, and (ii) downstream effectiveness in identifying anomalies. 

Using these measures, we compare a number of strong baselines and show that the proposed algorithm outperforms  oracle-knowledge policy sampling and the Gibbs-prior sampling approach suggested in \cite{grushka2019simulating}.

The novelty of our work lies in adding diversity as an important objective of monitoring and redefining the monitoring problem as a bandit problem, as well as in showing that a variant of a classic bandit sampling algorithm can improve existing approaches.

\section{Background and Motivation}
\subsection{Database Monitoring}

 Monitoring database and file activity enables organizations to issue alerts when dangerous events have occurred, and database monitoring is implemented in many domains, including health, finance, and insurance. Security information and event management (SIEM)  and DAM systems audit database activity and apply anomaly detection algorithms in an attempt to detect data misuse, data leakage, impersonation, and attacks on database systems. These systems are used by an organization's SO to identify suspicious activity \cite{kaplan2011meeting}. Unlike  recommendation systems where the users modeled are also the ones receiving the recommendation, in DAM settings the aim model the database users' activity in order to make recommendations for the SO regarding which users should be monitored.
 
When monitoring only users suspected of preforming risky activity without exploring and diversifying the monitored user list, the SO can only learn about the "normal behaviour" of a small group of users without getting a sense of what is normal for the majority of users and activities.

 \subsection{Transaction Risk}
 When an SO assigns risk to a transaction, various contextual information is used,  such as time of day, user activity profile, location (IP address), the nature of the activity (\textit{i.e.} is it permitted), data sensitivity, and the resulting data volume.
 
 When a DAM system is installed in an organization these rules can be defined  manually (by the SO) as a risk policy or learned by annotating risk scores on some representative transaction using a classifier such as CyberRank \cite{grushka2016cyberrank}. 
 
 This allows for the aggregation of the overall user activity and the assignment of a single risk-score for user activity within a time-frame. Methods for aggregating this are out of scope of this work, but simple aggregations such as maximum or median risk during a particular hour,  can be used.

\subsection{Increasing diversity when sampling for  monitoring}
The problem of data velocity and size when monitoring systems is not limited to the DAM domain, nor is the filter bubble effect. Studied conducted in the domain of network traffic flows such as \cite{jadidi2016intelligent,mai2006sampled} attempted to diversify the monitored data while maintaining the ability to detect malicious events. Mai \textit{et al.} \cite{mai2006sampled} found that for network flows the attempted sampling schemes weakened the anomaly detection ability to identify abrupt changes in volume - the sampling schemes degraded the performance examined in terms of success detection and false positive ratio. Kumar and Xu \cite{kumar2006sketch} suggested approximating the size of the traffic flow as a prior for the sampling probability, using count sketch for guiding the sampling method to estimate flow traffic measurement.

These studies found that  sampling only high-risk packets is deleterious for anomaly detection and that sampling introduced bias to the recalls of the anomaly detection.

\subsection{DAM Sampling as a Multi Armed Bandit Problem}

The problem of maximizing reward when sampling in the MAB problem has been extensively explored \cite{may2011simulation,agrawal2012analysis,chen2013combinatorial}. In multi-armed bandit problems, the optimal sampling (exploration / exploitation) strategy of playing against a slot machine with N arms each with an unknown reward distribution, is sought. 

In the data-base monitoring domain, the goal is to find the optimal strategy for sampling users' database transactions, using the available resources in order to maximize risk monitoring. This does not fit the classical MAB setting as we are pulling multiple levers in every round. The scenario of making multiple actions at each round with a constraint on budget had been recently described by \cite{zhou2018budget} as Budget-Constrained MAB with Multiple Plays. However, in our case the reward distribution for each arm may change during the game and the rewards are not sparse as some reward is gained from every armed pulled (unlike vanilla MAB problems where most actions result in no reward). We name this MAB variant Budget-Constrained-Dynamic-MAB.

Once we redefine the security sampling problem as a MAB problem we can define the sampling algorithms in MAB terms of exploration and exploitation and use reward as the objective, aligned to the task of maximizing the collection of informative and useful logs.

\section{Problem Setting}
We address the problem of diversifying the list of users who's transactions are being monitored based on the need of the system to monitor and capture all potential malicious activity, and the need of the SO to learn about the different users activity. Hence, we define the learning to diversify problem for database activity monitoring  along with maximization of risk "reward" subject to the available storage and computing capacities.
We now create a few notations to explain the problem. Let’s consider a policy $p_t$  at a certain time iteration $t \in \{ 0, ...., \infty \}  $ monitors  $S_t$ subset of the system users $ U = \{ u_1,...,u_n\}$ whos risk scores for time t are given by $ \{r_1,...r_n \}$. 
Our goal is to select the best policy $p_t$ (A subset of users from the n users) for each time-frame $t$ such that the result set caters to maximizing the  monitored risk score subject to the given capacity $C$. Let us use indicator variables $x_{jt} = \{0, 1\}$ denotes if the user $u_j$  is selected by policy $p_t$ to be monitored during time-frame t, and $z_{jt} = \{0, 1\}$ denotes if the user $u_j$  is selected by the oracle policy  $o_t$ to be monitored during time-frame t . The reward generated from the policy $p_t$ can be computed as $ \rho_t = \sum_{j=1}^{j=n}x_{jt}r_{jt}$. The reward proportion $r_t$ for time t can be computed by $$R_t = \frac{\rho_t}{\sum_{j=1}^{j=n}z_{jt}r_{jt}} $$  The total reward can then be 
given by $ R_T = \sum_{t=1}^{t=T}R_t$. Given the above set up learning to diversify problem in Database activity monitoring  can
be mapped to maximizing $R_T$ for a given capacity $C$ without degrading the Anomalies found. 

Each users activity is either fully monitored or fully discard.The anomaly detection aims to detect advanced persistent threat (APT) \cite{tankard2011advanced}. a single risk score for the users' time-frame activity lines up with these goals.

\section{Evaluation}
In this study, we use the following metrics to evaluate the performance of  sampling algorithms from the perspective of risk detection, coverage, and malicious events detected:
\begin{itemize}
    \item \textbf{Reward} 
    The main objective of the sampling strategy is to maximize the monitored risk . The oracle strategy,  detects the maximal risk $ \rho_{opt} = \sum_{j=1}^{j=n}z_{jt}r_{jt}$ for a given capacity  at time t.
    The reward for a time-frame is  the proportion of the detected risk out of the oracle strategy detected risk: 
    $R_t =\frac{\rho _t}{\rho_{oracle}}$
    \item \textbf{Coverage} 
    Anomaly detection systems rely on historical data in order to detect anomalies. Therefore acquiring risk profiles for as many users as fast as possible is highly important, also for avoiding the cold start problem. If we never sample the users suspected to be a low risk users the anomaly detection wouldn't be able to detect a change in their activity. Moreover, the SO perception of the user’s risk profile may be inaccurate or just missing. We measure how long does it take the system to collect data regarding the entire population. Therefore, we measure the percent of the population for which the sampling algorithm logged at least one time frame of activity. Let us define $users\_coverage_{t_i}= \{U_t \mid t \in \{1,...,t_i \} \}$ where $U_t$ is a list of all users monitored during time $t$
    \item \textbf{Recall of anomalies} 
    Downstream of the logged data, anomaly detection is applied to user activity. Previous work \cite{chandola2009anomaly} found that sampling adds bias to the results of the anomaly detection. 
    We measure the recall of an identical anomaly detection algorithm when applied to the different data sets produced by each sampling strategies. We normalize the recall to the recall discovered by applying the anomaly detection over the full data set (no sampling).

\end{itemize}
  
\section{Sampling algorithms}
We compare three algorithms for the problem. The first two baseline algorithms proposed in \cite{grushka2019simulating}. We introduce a novel algorithm named C\textendash	 \(\epsilon \)\textendash Greedy based on $\epsilon$-greedy   algorithm for the Budget-Constrained-Dynamic-MAB problem.

\subsection{SO-policy}
In each round, the C users with the highest known risk are sampled (these remain constant as no exploration is performed). Initialization is explained in the Experimental settings section.

\subsection{Gibbs-by-risk} 
In the Gibbs-exploration strategy, suggested by \cite{grushka2019simulating}, the user probability to be sampled is proportional to the risk estimate.

\begin{algorithm}[H]
\caption{Gibbs by risk}
\SetAlgoLined
\KwData{$U=\{u_1,..,u_n\}$, $R_{t-1}=\{r_{1,t-1},..,r_{n,t-1}\}$,
        $C>0$} 
\KwResult{list of users for time $t$ } 
 monitor\_users =\{\} 
 
\For{u in \{$u_1$,...,$u_n$\}}{
     $p_u$  = ${r_{u_i}}/{\sum_{i=1}^N r_{u_i}}$

 }
 
 \While{	len(monitor\textunderscore users) $\leq C$}{
          draw $u_i$ with probability $p_i \sim (p_1,...p_n)$

\If {$u_i \notin$ monitor\textunderscore users}{
             monitor\textunderscore users.append($u_i$)
        }
        
     }

\end{algorithm}


Sample randomly with a user's risk profile as the prior for choosing it. The user's probability to be sampled is determined by the maximal user$'$s risk found in the past (using a sliding window of k time frames) or the initialization value.

\subsection{C-$\epsilon$-greedy}
The \(\epsilon \)\textendash	greedy strategy for MAB has been shown to outperform other algorithms on most settings \cite{kuleshov2014algorithms}. Thompson sampling, another top performing algorithm for MAB, is not applicable to the non-discrete nature of Budget-Constrained-Dynamic-MAB.
In our setting we present C\textendash	 \(\epsilon \)\textendash Greedy strategy, 
where in each step the  \(\epsilon \times C\) (capacity) users with the highest risk are sampled, and \((1-\epsilon) \times C\) random users are sampled as well for exploration. 

\begin{algorithm}[H]
\SetAlgoLined
\KwData{$U=\{u_1,..,u_n\}$, $R_{i,t}=\{r_{1,1},..,r_{n,n}\}$,
        $C>0$, $k>0$} 
\KwResult{list of users for time t }
 
 monitor\_users =\{\}
 
 \For{$u_i \in U $}{
 $aggrisk_{u_i} = \sum_{t=T-k}^{T}r_{u_i,t}$
 
 $p_{u_i}$ = $aggrisk_{u_i}/k$

}
 
 sorted\textunderscore users $\gets sort(((u_1,p_1),..,(u_n,p_n)),desc)$
 
 $i = 0$
 
\While{$i< \epsilon \times C$}{
   monitor\textunderscore users.append(sorted\textunderscore users[i][0])
    
   $i+=1$
  
  }
 
 $i = 0$
 
  \While{$i< (1-\epsilon) \times C$}{
   draw j with probability $U \sim (1,..,n)$

   \If{sorted\textunderscore users[j][0] $\notin$ monitor\textunderscore users}
   {monitor\textunderscore users.append(sorted\textunderscore users[j][0])
   $i+=1$
   }
   
   }

 \caption{C-$\epsilon$-greedy}
\end{algorithm}


The SO-Policy can be viewed as a special case of the C\textendash$\epsilon$\textendash	Greedy Strategy where the exploitation is set to 100\%. 

\subsection{Random sampling} 
A baseline where users are sampled completely randomly at every time frame. This can be viewed as a special case of the C\textendash$\epsilon$\textendash	Greedy Strategy where the exploration is set to 100\%.

\section{Simulating DAM monitoring data}

Using data collected from an operating DAM system is irrelevant in our case  mainly because it contains bias introduced by the process of the data collection; Due to the high velocity nature of database activity, it is impossible to log all the activity, therefore the data collected using a sampling policy, hence contains bias. Other reasons we worked with simulated data: (1) The logged data may contain organizational Intellectual property (IP) and sensitive data and as such it is hard to convince commercial organization to share it. (2) Security events are not tagged in the logged data, therefore for each suspected activity a consultation with SO is needed. Verifying that the activity is malicious requires manual investigation of the activity and it's context making it impractical for research purposes. Using simulated data allows us to evaluate the effect of the sampling strategy using all the evaluation metrics described above. 

Collecting meaningful and unbiased data sets for studying effects of diversity and comparing sampling strategies is also a problem in other domains \cite{chen2019serendipity,lupu2019leveraging} such as search and recommendations, using simulated data instead allows for flexible and robust experimentation \cite{goswami2019learning,chen2019serendipity,germano2019few,voelkl2018reproducibility}. 

DAM systems log various features for each user transaction, using the data as is makes evaluating log coverage and  anomaly detection more complex as it is not a trivial task. To reduce the complexity of the features for comparison and evaluation, working with aggregated data is useful. Previous work such as \cite{grushka2016cyberrank,evina2019enforcing} leveraged SO knowledge to aggregate database activity into a single risk score.  \cite{grushka2019simulating} suggested a simulation package made of low complexity data where the user activity for a time frame is represented by a single aggregated risk score. Using simulated data also helps us with avoiding the bias introduced to the data set during the collecting stage and the ability to introduce anomalies in a controlled fashion.

The simulation is designed to mimic the properties of real users whose behavior is not random: users exhibit power log distribution of risk profile, few users produce most of the risky transactions, and transactions per user over time have a trend (see \cite{grushka2019simulating}). Before implementing our experiment we consulted a security expert to confirm that the simulated data matches the expected behaviour as observed in real data-bases activity.

\begin{figure}[h]
        \centering
\includegraphics[width=\linewidth]{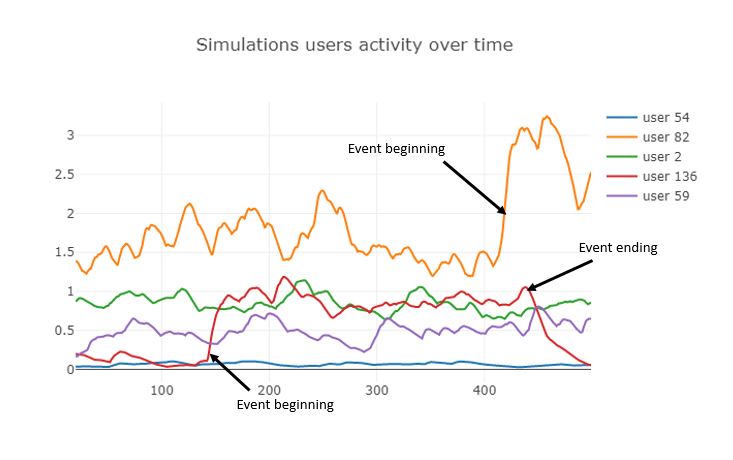}

  \caption{Risk distribution. Risk sampled per user over time produced by the simulation, two of the users have an anomaly introduced}
  \label{users_activity_over_time}
  \end{figure}

\section{Experimental settings}
 We used the simulation to produce 10 data sets using different random seeds and report on the average results. Each data set simulates activity profiles of 200 users for 3,000 time frames.
 The capacity (C) was set to 10\% of the numbers of users. 
 
 \subsection{Initialization}
 The setup of a DAM system relies on the SO's acquaintance with the databases users as well as his/her familiarity with  the domain potential risks. During the setup process the SO defines a monitoring policy made of  rules and user groups which represent the users and activities suspected of being the main threats to the organization's data. To examine the effect of that knowledge over the quality of the monitoring and anomaly detection and to assess whether the results of the SO's manual efforts,  can be leveraged as a prior for the cold start scenario we compare two settings: 

(1) A setting in which there is \textit{oracle initial knowledge} - This setting assumes that the SO, defining the policy, has perfect knowledge. To mimic this knowledge we sample the risk for each user at time $t_0$ and use it as the risk prior. (2) A setting in which there is a noisy oracle - this settings assumes that the SO has imperfect knowledge about the users' activity and the risk they present to the organization. In such cases, we use the risk from time $t_0$ mixed with noise from another randomly generated user (risk distribution) as a risk prior for each user.

\subsection{Introducing security events}
The anomaly detection component of DAM systems relies on modeling users' activity and detecting activities that are incompatible with that distribution. To evaluate the effect of the sampling strategy  on the anomaly detection component, we simulated security events. A security event, in which a user has been compromised or abused his/her permissions, is continuous (lasts more than a single time frame) and characterized by a change in the user's risk distribution. 

To simulate such an event we change the user's risk distribution, sampling a new risk distribution randomly for a period length $l$ which is sampled uniformly between 200 and 300. In each time frame, a user has a probability of experiencing such an event (set at 0.001 for these experiments).

Figure \ref{users_activity_over_time}, presents the activity of five different users from the users generated by the simulator. Each line indicates the users' hourly activity risk score. The graph presents one high-risk profiled user (user 82), two medium-risk profiled users (users 2 and 59), and two low-risk profiled users (users 136 and 54). User 82 experienced a security event beginning at the 414 time frame. User 136 experienced a security event between the 143 to  441 time frames. In the figure it can be seen that user 143 started of as a low risk user, this user’s activity profile changed to a medium-risk profile during the security event, while user 82 had a high-risk profile to begin with and the security event exacerbated this, resulting in an even higher-risk profile for this user. 

\subsection{Controlling diversity}
To evaluate the effect of different amounts of introduced diversity we tweak the $\epsilon$ part of the  C\textendash	 \(\epsilon \)\textendash Greedy algorithm. When $\epsilon$ is 1 only the highest risk users are sampled and no exploration is happening, this is essentially the SO-policy baseline. Setting $\epsilon$ to 0 means the system is in pure exploration mode, sampling users at random. 3 other settings are evaluated: setting $\epsilon$ to 0.2 (sample 80\% of capacity at random), setting $\epsilon$ to 0.5 (sample 50\% of capacity at random and 50\% from the highest risk users) and setting $\epsilon$ to 0.8 (sample only  20\% of capacity at random),

\section{Results}
\begin{table*}
  \caption{Reward and Recall for sampling strategies}
  \label{tab:rewards_and_recall}
  \begin{tabular}{ccll} 
    Strategy & Reward (detected risk, Oracle SO) & Reward (with Noisy-Oracle SO)  & Recall (detected anomalies)\\
    SO policy  & 0.677 & 0.514 & 0.084 \\ 
    Random   & 0.264 & 0.264 & \textbf{0.878}\\ 
    Gibbs by Risk & 0.514 & 0.39 & 0.56\\ 
    C\textendash	 \(\epsilon \)\textendash Greedy 0.2 & 0.479 & 0.48 & 0.834\\
    C\textendash	 \(\epsilon \)\textendash Greedy 0.5 & 0.703  & 0.702 & 0.779\\ 
    C\textendash	 \(\epsilon \)\textendash Greedy 0.8 & \textbf{0.864} & \textbf{0.863} & 0.67\\ 

  \end{tabular}
\end{table*}

\subsection{Performance}

Table \ref{tab:rewards_and_recall} presents the performance for the reward and recall metrics when using the various sampling strategies. As can be seen, the C\textendash	 \(\epsilon \)\textendash Greedy sampling strategy, with exploration set at 20\% or 50\% ($\epsilon$ set at 80\% or 50\%)  out-performed all of the other strategies. C\textendash	 \(\epsilon \)\textendash Greedy, with exploration of 20\% ($\epsilon$ set at 80\%) performed the best in terms of reward, with reward of 86\%. when initialized With a perfect SO knowledge (assuming the SO has the right estimation of all the users activity during setup) the SO-Policy of sampling only the riskiest users achieved a 67\% of the optimal reward, however when the initialization is noisy (imperfect expert knowledge at system setup) the result degraded to 51\% of the reward while the C\textendash	 \(\epsilon \)\textendash Greedy methods did not degrade significantly.
\begin{figure}
  \centering
\includegraphics[width=\linewidth]{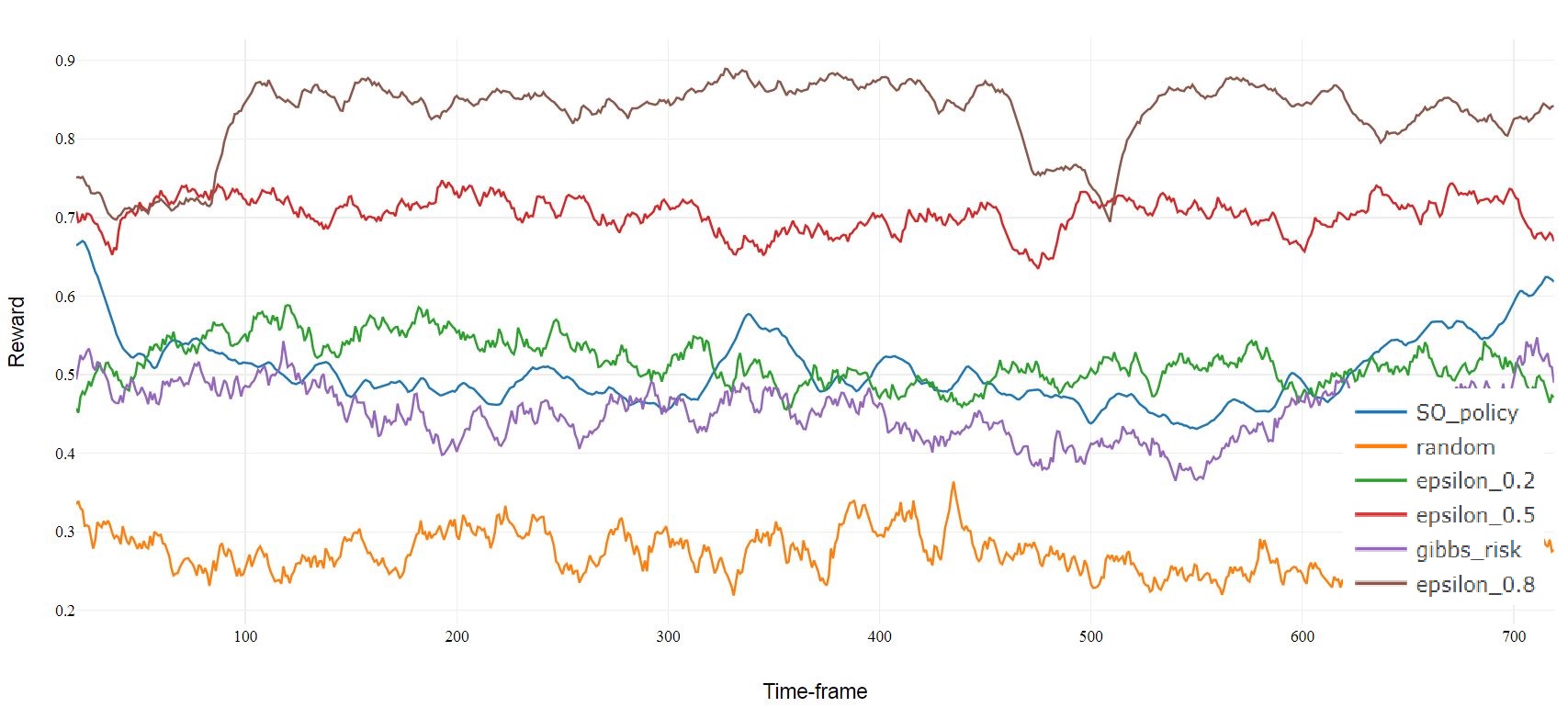}
  \caption{Risk Reward per Time Frame}
  \label{fig:Risk Regret Comparison}
\end{figure}
Figure \ref{fig:Risk Regret Comparison}, which describes the reward achieved in each time frame, shows that the C\textendash	 \(\epsilon \)\textendash Greedy sampling strategy both with the  80\% / \%20 exploitation / exploration ratio and 50\% / 50\% exploitation / exploration ratio, outperforms all other strategies, while random sampling constantly results in the poorest performance. In the figure, we see that  C\textendash	 \(\epsilon \)\textendash Greedy 80\% (using 20\% random exploration) takes about 100 time frames to explore, discovering the most risky users and then exploiting gaining ~20\% better reward than  C\textendash	 \(\epsilon \)\textendash Greedy 50\%. At time frame 460 we see another event where a change in the user population, a non-risky user affected by an anomaly that made it emit high risk transactions, cause both  C\textendash	 \(\epsilon \)\textendash Greedy 80\% and 50\% to experience a drop in reward, the reward goes back up once the offending user is discovered and added to the exploitation list. 


In terms of anomalies recall, the SO-knowledge policy detected only 8.5\% of the anomalies, not surprising as only 10\% of the users are ever monitored. The highest recall was found for random sampling detected 88\% of the anomalies. The C\textendash	 \(\epsilon \)\textendash Greedy sampling strategy detected between 67\% to 83\% of the anomalies. Gibbs-by-risk sampling got significantly less recall with 56\%.

\subsection{Coverage experiment} 
\begin{figure}[H]
  \centering
 \includegraphics[width=\linewidth]{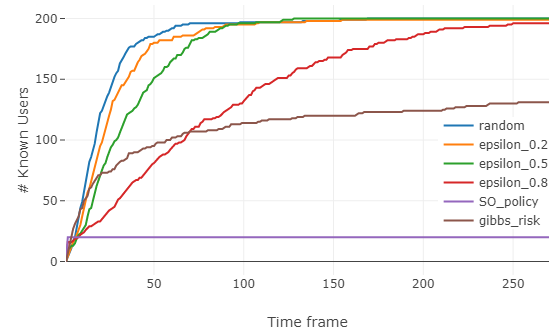}
  \caption{Knowledge acquisition rate}
  \label{fig:User information over time}
\end{figure}

In figure \ref{fig:User information over time} we compare the coverage of the different algorithms. On the x axis we shows the time and the Y axis shows the corresponding coverage. 

We marked a user as "covered" when the method gathered two samples of the user.

The SO-policy always samples the same group of users, therefore it is constant from the beginning as no other users are explored. With Gibbs-by-risk method there's a probability for any user to be sampled in proportion to their risk. We observe that users who exhibit low risk in initialization are not likely to be sampled which slows down the exploration of all users (not all users were sampled even after 3,000 time frames).

The strategies preferring completely random exploration had faster knowledge acquisition, depending on the proportion of the sample capacity devoted to exploration. When sampling with a 100\% exploration rate (completely random), 80\% exploration (epsilon\_0.2), or 50\% exploration the strategy had gathered  two samples for most users  by the 100th time frame. Dedicating 20\% of the capacity to exploration (epsilon\_0.8) requires $\sim250$ time frames for the same amount of  knowledge acquisition.

\begin{figure}[H]
  \centering
 \includegraphics[width=\linewidth]{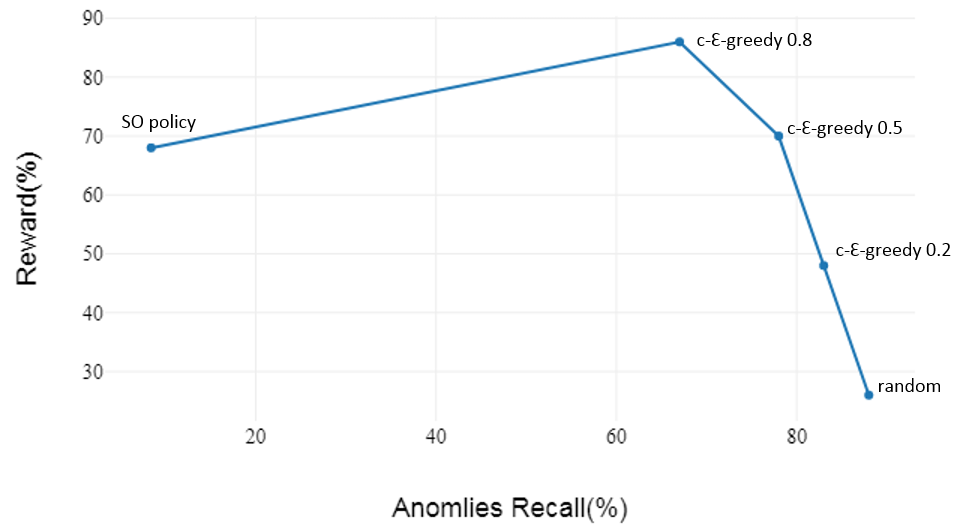}
  \caption{Reward vs Recall as a function of exploration rate}
  \label{fig:reward_vs_recall_as_func_of_exploration_dam}
\end{figure}
\subsection{Anomaly detection recall vs. reward}

Figure \ref{fig:reward_vs_recall_as_func_of_exploration_dam} shows that anomalies recall improves as we increase the exploration rate. However, the highest exploration rate yields low reward.

\section{Conclusions}

In this paper, we explored the effects of diversity when sampling activity for monitoring. We formulated the problem of sampling transactions in the security domain as a MAB problem, introducing Budget-Constrained-Dynamic-MAB, a multi-armed bandit where in each time-frame we sample a number of arms defined by the capacity of the system. As a user's risk profile may change when they switch position or become compromised, the reward probability may change during the user's life cycle. We extended the \(\epsilon \)\textendash Greedy algorithm, one of the best performing MAB approaches \cite{chen2013combinatorial}, to Budget-Constrained-Dynamic-MAB. Our variant, C\textendash	 \(\epsilon \)\textendash Greedy splits the sampling capacity at each turn into exploration part and exploitation part.

Using simulated data sets of user activity we showed that ensuring diversity when sampling database activity is important for understanding user behavior (coverage) and enhances downstream anomaly detection without adverse effects from allocating some capacity to diversity. 

The C\textendash	 \(\epsilon \)\textendash Greedy strategy that used 20\% of the capacity for exploration was able to beat a strong oracle-policy baseline on collecting risky activity, while achieving good coverage of all the population. The baseline method of a constant policy did achieve fine results on reward, collecting logs of risky activity, but was sensitive to initialization (depending on the SO familiarity with all the users). C\textendash	 \(\epsilon \)\textendash Greedy having better coverage allowed it to identify users becoming riskier, achieving better reward overall and was immune to bad initialization. 

We found that recall is driven by the exploration part of the strategy, and that completely random exploration had the best recall but worst reward (most of the logging capacity was spent on non-risky activity). 

Therefore, we advise using methods for sampling that allow the SO to dedicate some capacity for exploration and diversity. We  find that the The C\textendash	 \(\epsilon \)\textendash Greedy strategy is easy to implement, the $\epsilon$ is an easy knob to turn for enhancing exploration when needed, it supports initialization with the SO knowledge avoiding cold start and keeping humans in the loop while maximizing the audit trail for all users and supporting robust anomaly detection downstream.

\bibliographystyle{aaai}
\bibliography{mybib}

\begin{thebibliography}{}

\bibitem[\protect\citeauthoryear{Agrawal and Goyal}{2012}]{agrawal2012analysis}
Agrawal, S., and Goyal, N.
\newblock 2012.
\newblock Analysis of thompson sampling for the multi-armed bandit problem.
\newblock In {\em Conference on Learning Theory},  39--1.

\bibitem[\protect\citeauthoryear{Auer, Cesa-Bianchi, and
  Fischer}{2002}]{auer2002finite}
Auer, P.; Cesa-Bianchi, N.; and Fischer, P.
\newblock 2002.
\newblock Finite-time analysis of the multiarmed bandit problem.
\newblock {\em Machine learning} 47(2-3):235--256.

\bibitem[\protect\citeauthoryear{Chandola, Banerjee, and
  Kumar}{2009}]{chandola2009anomaly}
Chandola, V.; Banerjee, A.; and Kumar, V.
\newblock 2009.
\newblock Anomaly detection: A survey.
\newblock {\em ACM computing surveys (CSUR)} 41(3):15.

\bibitem[\protect\citeauthoryear{Chen \bgroup et al\mbox.\egroup
  }{2019}]{chen2019serendipity}
Chen, L.; Yang, Y.; Wang, N.; Yang, K.; and Yuan, Q.
\newblock 2019.
\newblock How serendipity improves user satisfaction with recommendations? a
  large-scale user evaluation.
\newblock In {\em The World Wide Web Conference},  240--250.
\newblock ACM.

\bibitem[\protect\citeauthoryear{Chen, Wang, and
  Yuan}{2013}]{chen2013combinatorial}
Chen, W.; Wang, Y.; and Yuan, Y.
\newblock 2013.
\newblock Combinatorial multi-armed bandit: General framework and applications.
\newblock In {\em International Conference on Machine Learning},  151--159.

\bibitem[\protect\citeauthoryear{Evina \bgroup et al\mbox.\egroup
  }{2019}]{evina2019enforcing}
Evina, P.~A.; AYACHI, F.~L.; JAIDI, F.; and BOUHOULA, A.
\newblock 2019.
\newblock Enforcing a risk assessment approach in access control policies
  management: Analysis, correlation study and model enhancement.
\newblock In {\em 2019 15th International Wireless Communications \& Mobile
  Computing Conference (IWCMC)},  1866--1871.
\newblock IEEE.

\bibitem[\protect\citeauthoryear{Germano, G{\'o}mez, and
  Le~Mens}{2019}]{germano2019few}
Germano, F.; G{\'o}mez, V.; and Le~Mens, G.
\newblock 2019.
\newblock The few-get-richer: a surprising consequence of popularity-based
  rankings?
\newblock In {\em The World Wide Web Conference},  2764--2770.
\newblock ACM.

\bibitem[\protect\citeauthoryear{Goswami, Zhai, and
  Mohapatra}{2019}]{goswami2019learning}
Goswami, A.; Zhai, C.; and Mohapatra, P.
\newblock 2019.
\newblock Learning to diversify for e-commerce search with multi-armed bandit.

\bibitem[\protect\citeauthoryear{Grushka-Cohen \bgroup et al\mbox.\egroup
  }{2016}]{grushka2016cyberrank}
Grushka-Cohen, H.; Sofer, O.; Biller, O.; Shapira, B.; and Rokach, L.
\newblock 2016.
\newblock Cyberrank: Knowledge elicitation for risk assessment of database
  security.
\newblock  2009--2012.
\newblock ACM.

\bibitem[\protect\citeauthoryear{Grushka-Cohen \bgroup et al\mbox.\egroup
  }{2017}]{grushka2017sampling}
Grushka-Cohen, H.; Sofer, O.; Biller, O.; Dymshits, M.; Rokach, L.; and
  Shapira, B.
\newblock 2017.
\newblock Sampling high throughput data for anomaly detection of data-base
  activity.
\newblock {\em arXiv preprint arXiv:1708.04278}.

\bibitem[\protect\citeauthoryear{Grushka-Cohen \bgroup et al\mbox.\egroup
  }{2019}]{grushka2019simulating}
Grushka-Cohen, H.; Biller, O.; Sofer, O.; Rokach, L.; and Shapira, B.
\newblock 2019.
\newblock Simulating user activity for assessing effect of sampling on db
  activity monitoring anomaly detection.
\newblock In {\em Policy-Based Autonomic Data Governance}. Springer.
\newblock  82--90.

\bibitem[\protect\citeauthoryear{Jadidi \bgroup et al\mbox.\egroup
  }{2015}]{jadidi2015performance}
Jadidi, Z.; Muthukkumarasamy, V.; Sithirasenan, E.; and Singh, K.
\newblock 2015.
\newblock Performance of flow-based anomaly detection in sampled traffic.
\newblock {\em Journal of Networks} 10(9):512.

\bibitem[\protect\citeauthoryear{Jadidi \bgroup et al\mbox.\egroup
  }{2016}]{jadidi2016intelligent}
Jadidi, Z.; Muthukkumarasamy, V.; Sithirasenan, E.; and Singh, K.
\newblock 2016.
\newblock Intelligent sampling using an optimized neural network.
\newblock {\em Journal of Networks} 11(01):16--27.

\bibitem[\protect\citeauthoryear{Juba \bgroup et al\mbox.\egroup
  }{2015}]{juba2015principled}
Juba, B.; Musco, C.; Long, F.; Sidiroglou-Douskos, S.; and Rinard, M.~C.
\newblock 2015.
\newblock Principled sampling for anomaly detection.
\newblock In {\em NDSS}.

\bibitem[\protect\citeauthoryear{Kaplan, Sharma, and
  Weinberg}{2011}]{kaplan2011meeting}
Kaplan, J.; Sharma, S.; and Weinberg, A.
\newblock 2011.
\newblock Meeting the cybersecurity challenge.
\newblock {\em Digit. McKinsey Google Scholar}.

\bibitem[\protect\citeauthoryear{Kuleshov and
  Precup}{2014}]{kuleshov2014algorithms}
Kuleshov, V., and Precup, D.
\newblock 2014.
\newblock Algorithms for multi-armed bandit problems.
\newblock {\em arXiv preprint arXiv:1402.6028}.

\bibitem[\protect\citeauthoryear{Kumar and Xu}{2006}]{kumar2006sketch}
Kumar, A., and Xu, J.~J.
\newblock 2006.
\newblock Sketch guided sampling-using on-line estimates of flow size for
  adaptive data collection.
\newblock In {\em Infocom}.

\bibitem[\protect\citeauthoryear{Lupu, Durand, and
  Precup}{2019}]{lupu2019leveraging}
Lupu, A.; Durand, A.; and Precup, D.
\newblock 2019.
\newblock Leveraging observations in bandits: Between risks and benefits.

\bibitem[\protect\citeauthoryear{Mai \bgroup et al\mbox.\egroup
  }{2006}]{mai2006sampled}
Mai, J.; Chuah, C.-N.; Sridharan, A.; Ye, T.; and Zang, H.
\newblock 2006.
\newblock Is sampled data sufficient for anomaly detection?
\newblock In {\em Proceedings of the 6th ACM SIGCOMM conference on Internet
  measurement},  165--176.
\newblock ACM.

\bibitem[\protect\citeauthoryear{Matt \bgroup et al\mbox.\egroup
  }{2014}]{Matt2014Escaping}
Matt, C.; Benlian, A.; Hess, T.; and Weiß, C.
\newblock 2014.
\newblock Escaping from the filter bubble? the effects of novelty and
  serendipity on users’ evaluations of online recommendations.

\bibitem[\protect\citeauthoryear{May and Leslie}{2011}]{may2011simulation}
May, B.~C., and Leslie, D.~S.
\newblock 2011.
\newblock Simulation studies in optimistic bayesian sampling in
  contextual-bandit problems.
\newblock {\em Statistics Group, Department of Mathematics, University of
  Bristol} 11:02.

\bibitem[\protect\citeauthoryear{Nguyen \bgroup et al\mbox.\egroup
  }{2014}]{nguyen2014exploring}
Nguyen, T.~T.; Hui, P.-M.; Harper, F.~M.; Terveen, L.; and Konstan, J.~A.
\newblock 2014.
\newblock Exploring the filter bubble: the effect of using recommender systems
  on content diversity.
\newblock In {\em Proceedings of the 23rd international conference on World
  wide web},  677--686.
\newblock ACM.

\bibitem[\protect\citeauthoryear{Tankard}{2011}]{tankard2011advanced}
Tankard, C.
\newblock 2011.
\newblock Advanced persistent threats and how to monitor and deter them.
\newblock {\em Network security} 2011(8):16--19.

\bibitem[\protect\citeauthoryear{Voelkl \bgroup et al\mbox.\egroup
  }{2018}]{voelkl2018reproducibility}
Voelkl, B.; Vogt, L.; Sena, E.~S.; and W{\"u}rbel, H.
\newblock 2018.
\newblock Reproducibility of preclinical animal research improves with
  heterogeneity of study samples.
\newblock {\em PLoS biology} 16(2):e2003693.

\bibitem[\protect\citeauthoryear{Zhou and Tomlin}{2018}]{zhou2018budget}
Zhou, D.~P., and Tomlin, C.~J.
\newblock 2018.
\newblock Budget-constrained multi-armed bandits with multiple plays.
\newblock In {\em Thirty-Second AAAI Conference on Artificial Intelligence}.

\end{thebibliography}
\end{document}